\documentclass[lettersize,journal]{IEEEtran}
\usepackage{amsmath,amsfonts}
\usepackage{algorithmic}
\usepackage{algorithm}
\usepackage{array}
\usepackage[caption=false,font=normalsize,labelfont=sf,textfont=sf]{subfig}
\usepackage{textcomp}
\usepackage{stfloats}
\usepackage{url}
\usepackage{verbatim}
\usepackage{graphicx}
\usepackage{cite}
\begin{document}

\title{Synthetic Face Ageing:  Evaluation, Analysis and Facilitation of Age-Robust Facial Recognition Algorithms}

\author{Wang Yao, 
Muhammad Ali Farooq, 
Joseph Lemley,~\IEEEmembership{Member,~IEEE,}
Peter Corcoran,~\IEEEmembership{Fellow,~IEEE}

\thanks{This work is funded by Irish Research Council Enterprise Partnership Ph.D. Scheme under Grant EPSPG/2020/40 and FotoNation, Ireland. }
\thanks{Wang Yao, Muhammad Ali Farooq and Peter Corcoran are with the College of Science and Engineering, University of Galway, Galway, H91 TK33 Ireland.}
\thanks{Joseph Lemley is with FotoNation, Galway, H91 V0TX Ireland.}
}

\markboth{THIS WORK HAS BEEN SUBMITTED TO THE IEEE FOR POSSIBLE PUBLICATION}%
{Shell \MakeLowercase{\textit{et al.}}: A Sample Article Using IEEEtran.cls for IEEE Journals}

\IEEEpubid{0000--0000/00\$00.00~\copyright~2021 IEEE}

\maketitle

\begin{abstract}
The ability to accurately recognize an individual's face with respect to human aging factor holds significant importance for various private as well as government sectors such as customs and public security bureaus, passport office, and national database systems. Therefore, developing a robust age-invariant face recognition system is of crucial importance to address the challenges posed by ageing and maintain the reliability and accuracy of facial recognition technology. In this research work, the focus is  to explore the feasibility of utilizing synthetic ageing data to improve the robustness of face recognition models that can eventually help in recognizing people at broader age intervals. To achieve this, we first design set of experiments to evaluate state-of-the-art synthetic ageing methods. In the next stage we explore the effect of age intervals on a current deep learning-based face recognition algorithm by using synthetic ageing data as well as real ageing data to perform rigorous training and validation. Moreover, these synthetic age data have been used in facilitating face recognition algorithms. Experimental results show that the recognition rate of the model trained on synthetic ageing images is 3.33\% higher than the results of the baseline model when tested on images with an age gap of 40 years, which prove the potential  of synthetic age data which has been quantified to enhance the performance of age-invariant face recognition systems.
\end{abstract}


\section{Introduction}\label{intro}
\IEEEPARstart{B}{iometric} characteristics including facial patterns, fingerprints and iris are unique to everyone. Developing a robust biometric authentication system by adopting these individual biometrics provides convenience and security in our daily lives. Biometric authentication systems such as doorbell cameras, smartphones and financial security systems have been used in everyday scenarios. Ensuring the longevity and security of biometrics is necessary for their broad adoption. However, it is noteworthy to mention that personal biometric characteristics change as time goes by. This phenomenon raises issues for advanced electronic biometrics systems. Studies~\cite{9025674} have shown that face recognition systems are affected by these biometric changes. 

One potential solution entail prompting users to periodically update their bioinformatic information within the system over the course of time. A common use case is the need for immigration authorities to update passport photos every few years. However, it is not possible to update the bioinformatics in some critical such as tracking escaped criminals for a long time and finding the missing individual after several years. Thus, developing a robust age-invariant face recognition (AIFR) system that is insensitive to the longitude ageing effect is a unique challenge.

Ideally, to design a long-term effective biometric authentication system, it is necessary to find large-scale balanced datasets that have adequate face data of different age stages for training the deep learning networks. However, collecting data on a person from birth to old age is difficult. Commonly used age datasets either have a small number of subjects~\cite{993553, 8014984} or the same subject spans a small age range~\cite{1613043}. In addition, the images from the existing public dataset did not use the same camera/imaging system, thus the quality of the images is inconsistent as the quality of smartphone cameras improved significantly after 2007.

In recent years various studies~\cite{10.1007/978-3-030-58539-6_44, 10.1007/978-3-031-19775-8_34} have been published with a specific focus on developing synthetic ageing algorithms in order to generate photo-realistic ageing samples. Inspired by these promising results from these synthetic ageing algorithms. In this work, we want to explore how synthetic ageing data facilitates the computer vision algorithms i.e., improving the robustness of face recognition on the ageing factor. To achieve this, firstly, good photo-realistic ageing samples that better imitate the human characteristics change due to the ageing effect should be adopted in the experiment. In this case, the following key research questions are being explored in this work.

\begin{itemize}
    \item How good are Generative Adversarial Networks (GAN) techniques in simulating ageing accurately?
    \item How robust are facial authentication methods concerning the human ageing factor? How will face recognition algorithms perform for an individual subject as they grow older?
    \item Is it possible for us to develop an age-invariant face recognition model by employing synthetic ageing data?
\end{itemize}

\IEEEpubidadjcol

To answer these questions, three state-of-the-art synthetic ageing methods are employed in this work which include SAM~\cite{10.1145/3450626.3459805}, CUSP~\cite{10.1007/978-3-031-19787-1_32} and AgeTransGAN~\cite{10.1007/978-3-031-19775-8_34} to generate synthetic ageing samples. We have designed a series of experiments to evaluate the synthetic ageing samples from the perspective of age accuracy and identity preservation. In addition, we have done a complete study on the comparison of the long-term real-world ageing effect and synthetic ageing effect. Moreover, we explore the potential of utilizing synthetic ageing data to compensate for face recognition algorithm performance degradation due to age intervals.
The main contribution of this work is as follows.

\begin{itemize}
    \item This work evaluates the synthetic ageing data by conducting age accuracy and identity preservation experiments.
    \item We first designed a self-consistency framework to evaluate identity preservation on synthetic ageing data.
    \item This work analyzes the impact of face ageing intervals of real-world data and synthetic data on the performance of face recognition algorithms.
    \item Further exploring the utility of synthetic ageing data in tackling the problem of age-invariant face recognition.
\end{itemize}

The rest of the paper is structured as follows. Section~\ref{background} presents a literature review including the effect of facial ageing on face recognition models, age-invariant face recognition methods and synthetic ageing methods. Section~\ref{evaluation} evaluates the synthetic ageing techniques qualitatively and quantitatively. Then Section~\ref{robustness} compares the robustness of face recognition models on real-world ageing data and synthetic ageing data. In addition, Section~\ref{finetune} conducts experiments to explore the use of synthetic ageing data to facilitate the age-invariant face recognition models. Finally, the conclusion of this research and discussion of the potential future work are drawn in Section~\ref{conclusion}.

\section{Background}\label{background}

\subsection{Ageing Effect on Face Recognition}
Analyzing the effect of human facial ageing on the performance of face recognition systems is useful for authentication applications such as passport renewal after every five- and ten-year time span. Two ways are adopted in terms of the facial ageing effect: ageing groups and ageing intervals. The performance of face recognition across different age groups is reported by different researchers~\cite{6327355, 9093357, WU2019116}. Most of them agreed that older adults are easier to recognize while younger are harder to recognise. The effect of ageing intervals indicates that recognizing an individual after several years. It is shown that a large time lapse between two face images from the same identity can lead to false reject errors~\cite{4409069, 8014816}. A detailed survey of the effects of ageing can be found in~\cite{sawant2019age}.  

These studies are tested on real-world datasets~\cite{8014984, 993553}, which are adopted in different years with various devices such as different mobiles and cameras. In this work, we have focused on exploring the effect of face recognition on a large-scale real-world ageing dataset and synthetic ageing data, by which to illustrate the effectiveness of the synthetic ageing data. Similar research was done by Grimmer et al.~\cite{9897043}, where authors conducted an initial study on evaluating the impact of face ageing on a face recognition system by using two synthetic ageing methods SAM~\cite{10.1145/3450626.3459805} and InterfaceGAN~\cite{9157070} to generate synthetic ageing data. They adopt two face image quality assessment methods including FaceQnet v1 and SER-FIQ to evaluate the synthetic age data. Different from them, we design a broader and more comprehensive method to evaluate the synthetic ageing data. 

\subsection{Age-Invariant Face Recognition}
AIFR is proposed for solving identity faces with a large age gap, which is different from general face recognition methods. Most of the existing algorithms adopt discriminative or generative methods to achieve AIFR. The discriminative methods extract age-invariant robust features for face recognition tasks. Gong et al.~\cite{6751468} proposed a Hidden Factor Analysis (HFA) approach to factorize mixed features and get the age factor and age-independent factor for recognition. Hou et al.~\cite{9711408} presented an AIFR framework which disentangles identity and age into two parts by using mutual information minimization. Yan et al.~\cite{10.1145/3472810} proposed a Multi-feature Fusion and Decomposition (MFD) framework which can exact robust identity features and is not sensitive to the ageing process. 

The generative methods, on the other hand, generate the face of the target age and transform the problem into identifying two faces with the same age interval. A typical AIFR architecture proposed by Zhao et al.~\cite{9146699}, combines cross-age face generation and recognition simultaneously which achieved remarkable results. Huang et al.~\cite{9931965} proposed a multi-task learning framework based on this approach which adopts an identity conditional module to achieve face synthesis with identity preservation. 

To summarise, previous work has used generative models to generate target faces which will help to minimize the differences in biometric characteristics caused by the ageing factor. Different from previous work, our work not only keeps the original facial features, but further also uses synthetic data to enlarge the existing dataset and explores which group of specific synthetic age will benefit the face recognition models. This process can be easily integrated in common end-to-end recognition networks. 

\subsection{Age Transformation}
Simulating face ageing across various ages presents a formidable challenge. Benefitting from the rapid development in the domain of generative artificial intelligence in recent years, the quality of synthesized images has been significantly improved. To some extent, researchers have explored methods to generate faces across various age groups while preserving the identity. Or-El et al.~\cite{10.1007/978-3-030-58539-6_44} employed a multi-domain image-to-image generative adversarial network architecture named LATS, which could generate a full head portrait with texture and shape. Hsu et al.~\cite{10.1007/978-3-031-19775-8_34} propose a novel network for identity-preserving facial age transformation named AgeTransGAN and a rectification scheme for improving the usage of the metrics. Li et al.~\cite{electronics12112369} proposed a gender-preserving face ageing model GFAM which incorporates gender- and identity-preserving components for age imbalance data. 

Others have explored synthesising the faces with the target age, i.e. modelling the continuous ageing process. Yao et al.~\cite{9412383} proposed HRFAE which adopts an encoder-decoder architecture and translates images from 20 to 70 years old. Alaluf et al.~\cite{10.1145/3450626.3459805} presented age synthesis as an image-to-image translation task and provided fine-grained control on the target face by using a regression task. Gomez-Trenado et al.~\cite{10.1007/978-3-031-19787-1_32} introduce the Custom Structure Preservation module (CUSP) which can produce structural modifications while preserving relevant details. Recently, Chen et al.~\cite{chen2023face} proposed a diffusion-based face ageing editing method referred to as ‘FADING’, which shows the potential usage of large-scale diffusion models for generating face ageing data. 

In this work, the main focus is on evaluating the synthetic age data, by exploring the robustness of face recognition models on the synthetic ageing data and tackling the age-invariant face recognition model with synthetic ageing data, rather than proposing a method for generating the photo-realistic ageing data. To achieve this goal, we adopted three state-of-the-art synthetic ageing methods as baseline algorithms for generating photo-realistic ageing data with target age and provide a general experimental framework to answer the above questions.  

\section{Methodology}

\begin{figure*}[!thpb]
    \centering
    \includegraphics[width=0.8\linewidth]{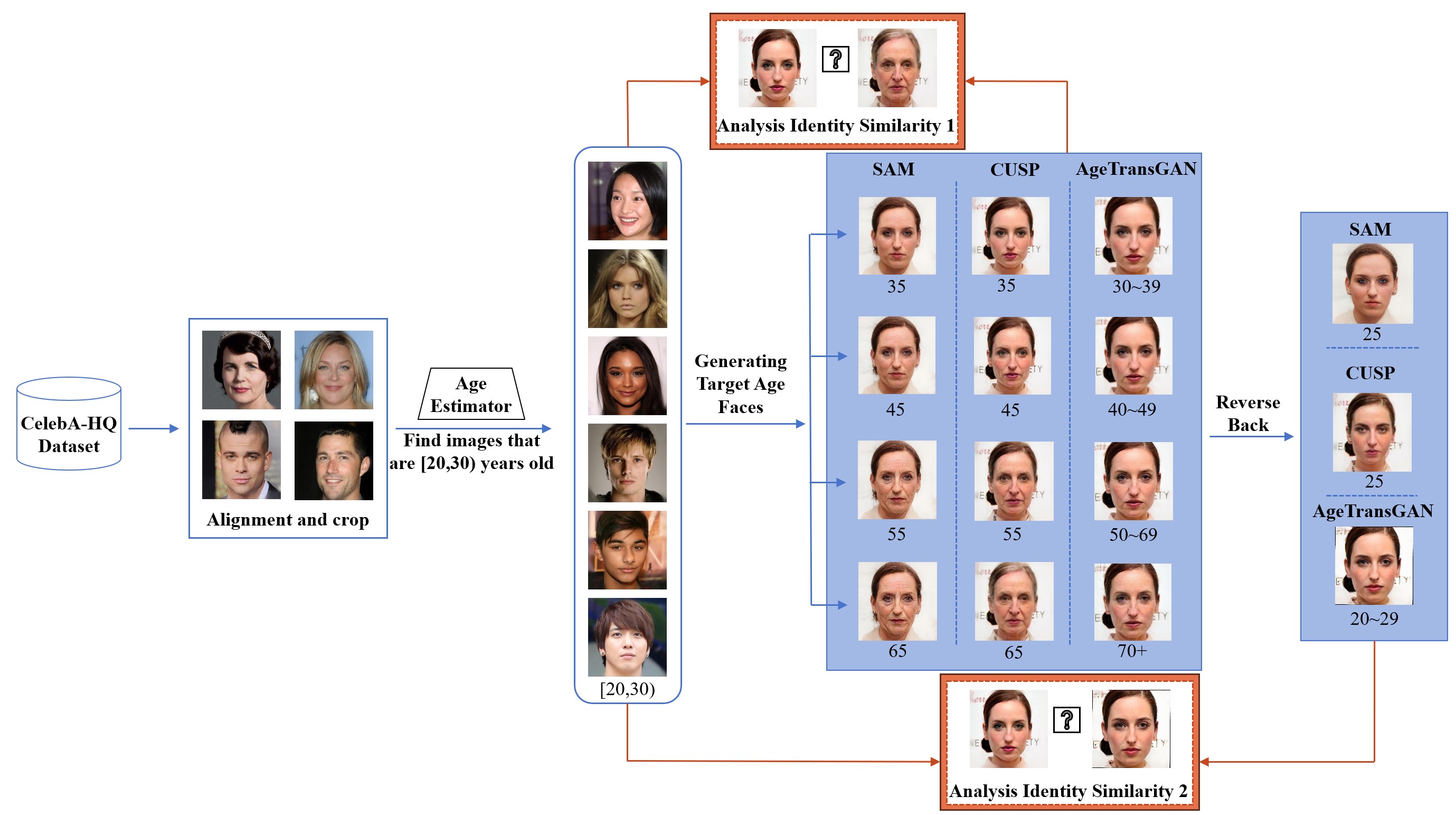}
    \caption{The process of identity preservation.}
    \label{process_id}
\end{figure*}

This section introduces the adapted methodology including synthetic ageing methods and related datasets. Firstly, we have generated synthetic ageing samples by using the following methods. Then, these samples are used for evaluation the age accuracy and identity preservation. After that, the effectiveness of these ageing samples is tested on face recognition algorithms. Finally, we have discussed whether we could use these synthetic ageing samples to improve the robustness of the face recognition algorithm. 

\subsection{Synthetic Ageing Methods}

Three state-of-the-art face ageing methods including SAM~\cite{10.1145/3450626.3459805}, CUSP~\cite{10.1007/978-3-031-19787-1_32}, and AgeTransGAN~\cite{10.1007/978-3-031-19775-8_34} are adopted in this work to generate photo-realistic ageing samples.  

\subsubsection{SAM}

The Style-based Age Manipulation (SAM) method~\cite{10.1145/3450626.3459805} approaches the task of age transformation as an image-to-image translation method. This process involves modifying an individual's appearance across their lifespan while preserving their underlying identity. Specifically, SAM operates by learning to encode target face images directly into the latent space of a pre-trained StyleGAN model, with a given aging shift. To achieve this, SAM employs a pre-trained age regression network to guide the encoder in generating latent codes that correspond to the desired age transformation. 

\subsubsection{CUSP}

CUstom Structure Preservation (CUSP) method~\cite{10.1007/978-3-031-19787-1_32} design a facial age editing approach that can generate structural modifications while preserving relevant details in the source face image.  This method enables users to adjust the extent of structure preservation in the input image during inference, which is different from the existing synthetic face age generation methods. To facilitate this functionality, this architecture introduces a masking mechanism termed the CUSP module, which discerns relevant regions in the input image from those to be discarded, all without necessitating additional supervision. 

\subsubsection{AgeTransGAN}

AgeTransGAN~\cite{10.1007/978-3-031-19775-8_34} comprises an encoder-decoder generator paired with a conditional multitask discriminator featuring an embedded age classifier, which is a novel framework for facial age editing. This method incorporates age features alongside the target age group label, collaborating with the embedded age classifier to ensure the generation of the desired target age. In addition, they propose rectified metrics for performance evaluation and evaluate AgeTransGAN against state-of-the-art approaches using both existing and rectified metrics. 

\subsection{Dataset}

In this work,  two different publicly available datasets are utilized, which include the CelebA-HQ dataset and B3FD dataset. The details of dataset attributes are as follows. 
\subsubsection{CelebA-HQ}
The CelebA-HQ dataset, a high-quality face dataset derived from CelebA~\cite{liu2015faceattributes}, comprises 30,000 images of $1024 \times 1024$ resolution faces featuring approximately 6,000 unique identities. This dataset was curated by extracting faces from the original dataset based on facial landmarks and enhancing image quality using a Generative Adversarial Network (GAN)-trained super-resolution model. To standardize image sizes, bilinear interpolation and frame filtering techniques were employed to scale images to $1024 \times 1024$ dimensions. CelebA-HQ stands out as a high-quality synthetic dataset characterized by an adequate number of subjects and limited lighting variations, rendering it suitable for face recognition tasks. In this work, we have used the CelebA-HQ dataset as a baseline to produce synthetic ageing data by using various face editing methods. 

\subsubsection{B3FD}
\begin{figure*}[!thpb]
  \centering
    \subfloat[SAM]{      
    \begin{minipage}[b]{0.3\linewidth}
		\includegraphics[width=\linewidth]{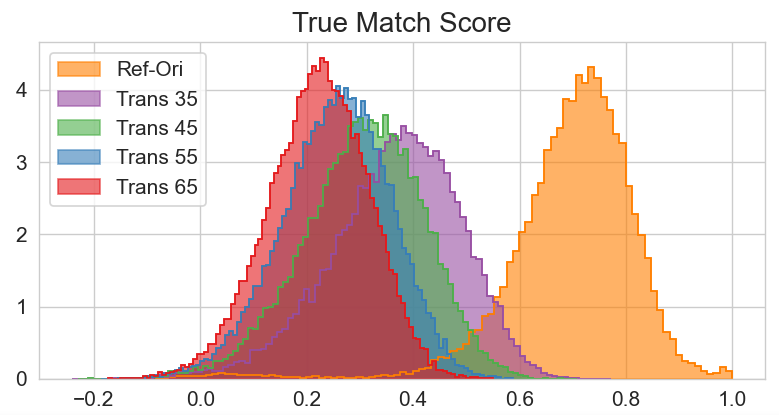} \\
		\includegraphics[width=\linewidth]{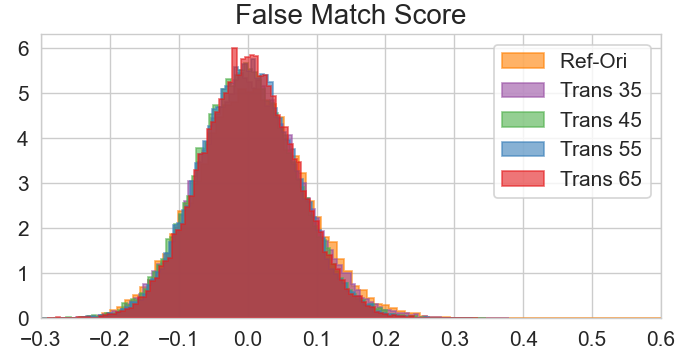}
    \end{minipage}
        }
    \subfloat[CUSP]{      
    \begin{minipage}[b]{0.3\linewidth}
		\includegraphics[width=\linewidth]{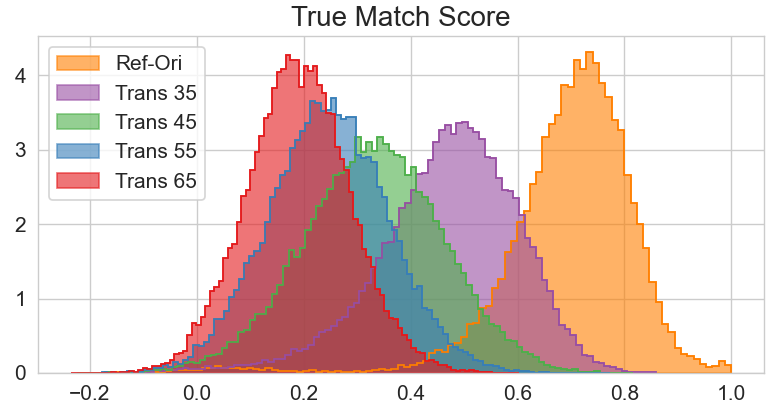} \\
		\includegraphics[width=\linewidth]{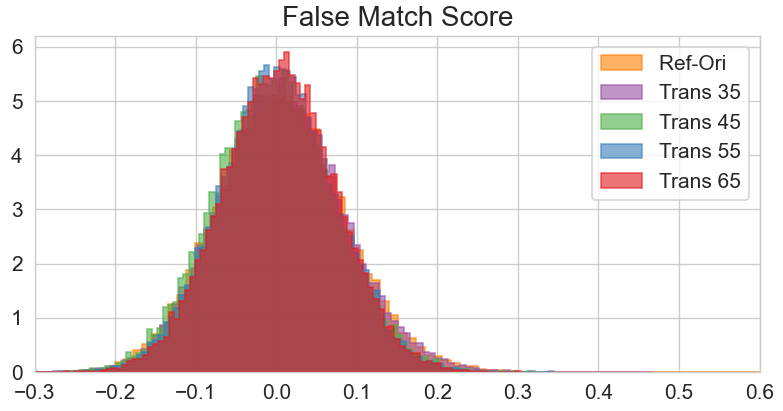}
    \end{minipage}
        }   
    \subfloat[AgeTransGAN]{      
    \begin{minipage}[b]{0.3\linewidth}
		\includegraphics[width=\linewidth]{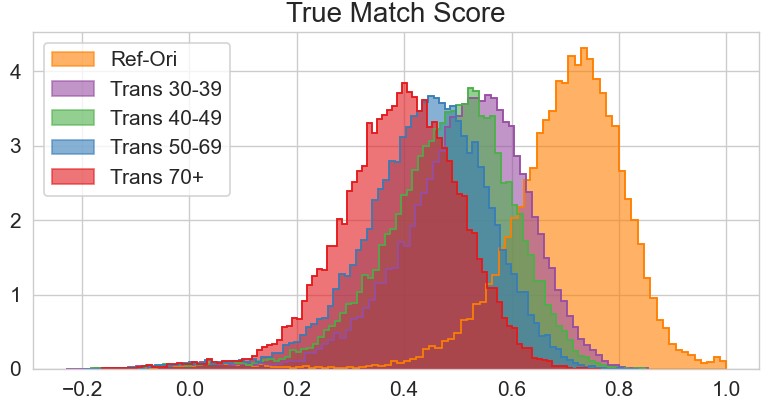} \\
		\includegraphics[width=\linewidth]{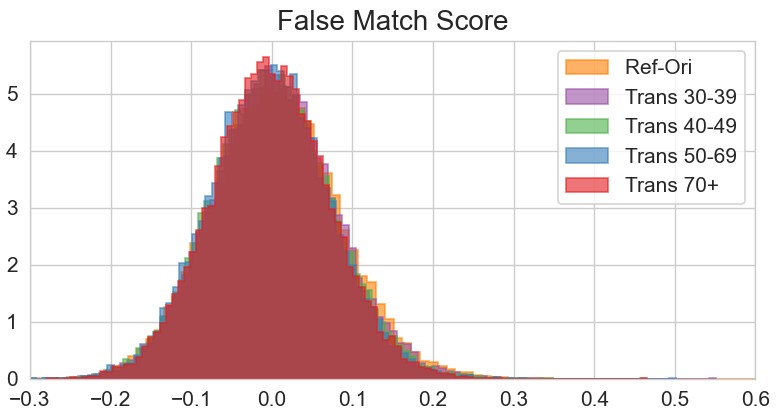}
    \end{minipage}
        } \\
    \subfloat[SAM]{      
    \begin{minipage}[b]{0.3\linewidth}
		\includegraphics[width=\linewidth]{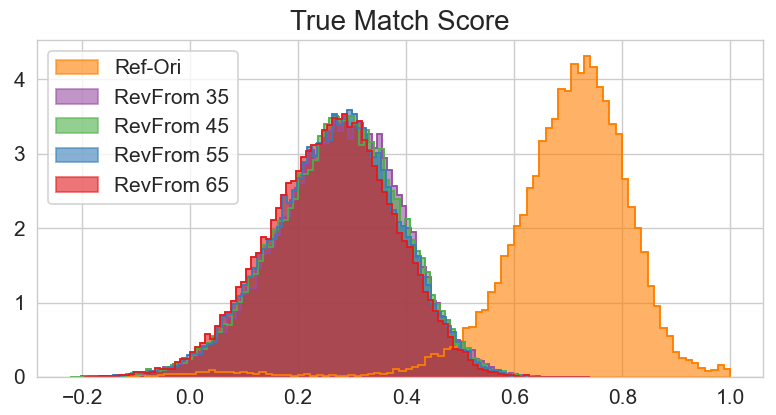} \\
		\includegraphics[width=\linewidth]{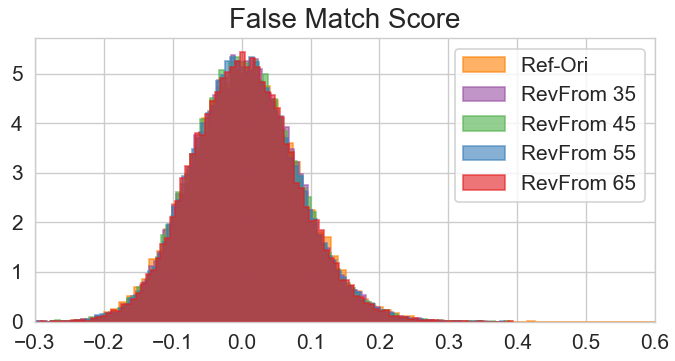}
    \end{minipage}
        }
    \subfloat[CUSP]{      
    \begin{minipage}[b]{0.3\linewidth}
		\includegraphics[width=\linewidth]{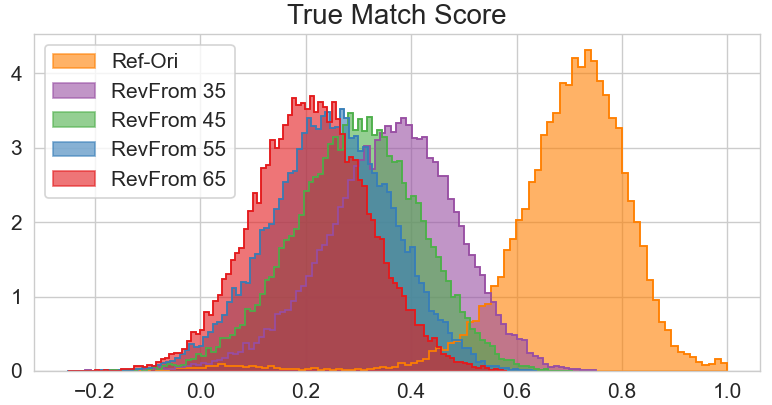} \\
		\includegraphics[width=\linewidth]{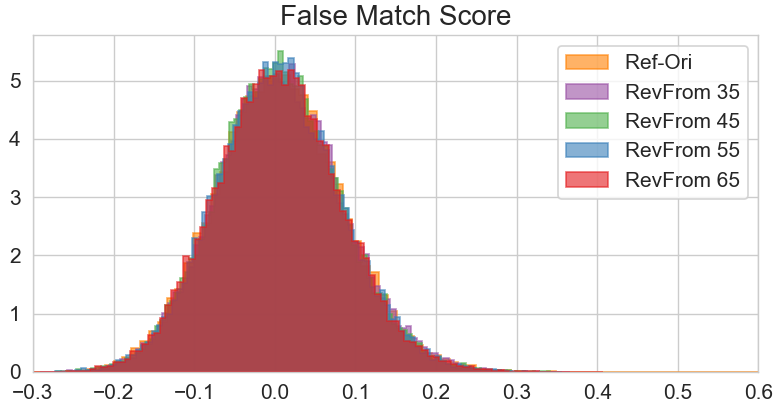}
    \end{minipage}
        }   
    \subfloat[AgeTransGAN]{      
    \begin{minipage}[b]{0.3\linewidth}
		\includegraphics[width=\linewidth]{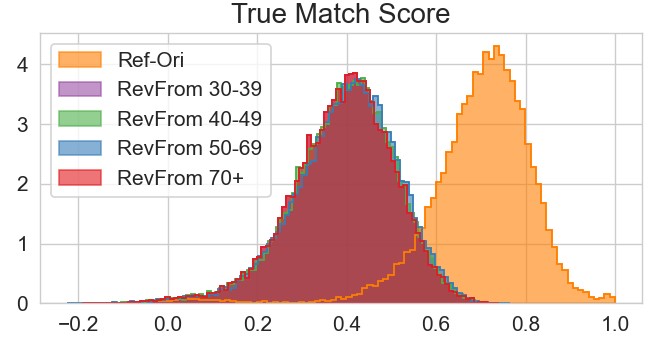} \\
		\includegraphics[width=\linewidth]{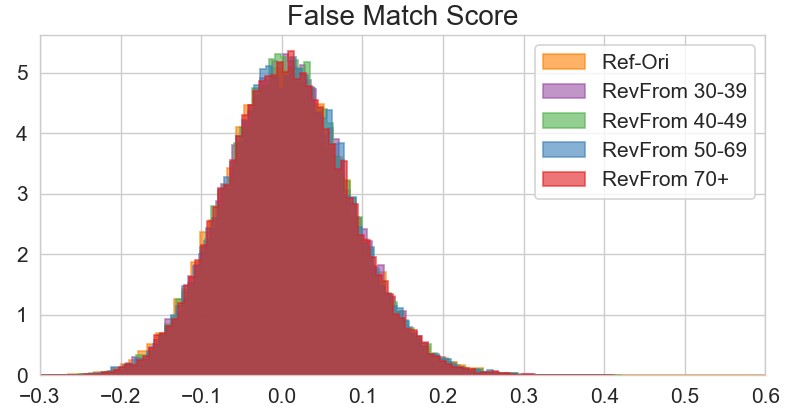}
    \end{minipage}
        }              
    \caption{Age identity analysis of the samples generated by SAM, CUSP, and AgeTransGAN methods. The top figure in (a)(b)(c)(d)(e)(f) shows the genuine distribution. The bottom figure in (a)(b)(c)(d)(e)(f) shows the impostor distribution. (a)(b) and (c) show the results of Analysis Identity Similarity 1 using the method from Figure~\ref{process_id}. (d)(e) and (f) show the results of Analysis Identity Similarity 2 using the method from Figure~\ref{process_id}.}
    \label{id_ana}
\end{figure*}
Biometrically Filtered Famous Figure Dataset~\cite{bevsenic2022picking} or B3FD has merged two large web-based datasets, including CACD~\cite{10.1007/978-3-319-10599-4_49} and IMDB-WIKI~\cite{Rothe2016DeepEO}. IMDB-WIKI dataset is the largest publicly available dataset of face images with gender and age labels,\footnote{\url{https://data.vision.ee.ethz.ch/cvl/rrothe/imdb-wiki/}.} which is collected directly from open internet sources. This suggests that the IMDB-WIKI dataset contains a significant amount of mislabeled data, despite its widespread utilization for training purposes. Considering that web-scraping approaches for automatic data collection can produce large amounts of weakly labelled and noisy data, B3FD is focused on cleaning the web-scraped facial datasets by automatically removing erroneous samples that impair their usability. B3FD has 375,592 images with 53,759 unique subjects. The age labels are ranging from 0 to 100. It comprises two subsets including B3FD-IWS and IMDB-WIKI. B3FD-IWS consists of 245,204 images from the IMDB-WIKI dataset with 53,568 unique subjects. B3FD-CS consists of 130,388 processed samples from the CACD dataset with 1,831 unique subjects. Since the data volume of this dataset is large enough, this dataset has become the baseline to explore the effect of real-world age intervals on face recognition. 

\section{Evaluation of the synthetic ageing samples}\label{evaluation}

In this section, we have evaluated the generated different ageing samples by age accuracy and identity preservation, which are two important parts for analysis of the synthetic ageing faces. Accurate age and identity are two important factors for generating photo-realistic faces with a target age. Many algorithms~\cite{9412383, 10.1007/978-3-031-19775-8_34, electronics12112369} utilizing GAN models for synthesizing age data incorporate both an identity preservation module and an age estimator module. Inspired by these algorithms, we evaluate the age accuracy and personal identity of the synthesized images. Other researchers have designed related experiments to assess these two factors. For instance, Liu et al.~\cite{Liu_2019_CVPR} employed Face++ APIs to estimate age and analyzed the distribution of estimated ages to evaluate age accuracy. To test identity preservation, they utilized verification confidence and accuracy metrics. In our work, we introduce two age estimators. One of them, DEX~\cite{Rothe_2015_ICCV_Workshops}, is a popular age estimator that could assess apparent age from a single image, and the other estimator, recently proposed by Hsu et al.~\cite{10.1007/978-3-031-19775-8_34} rectifies the bias of the Face++APIs estimator. We used box diagrams to visualize and analyze the difference between the target age and the assessed age. Regarding to identity preservation, we designed turn-old and reverse-back experiments to analyze changes in identity in two ways. By adding backtracking experiments, we can learn about the consistency of the identity during the process of generating the rest of the ages again using the synthetic age data. 

\subsection{Identity Preservation}

\begin{figure*}[!thpb]
    \centering
    \includegraphics[width=0.32\linewidth]{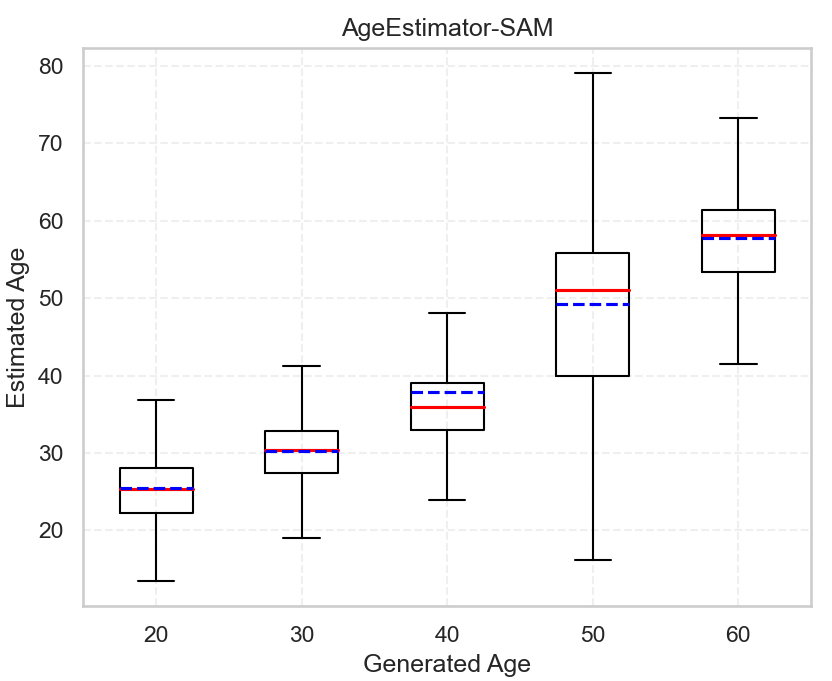}
    \includegraphics[width=0.32\linewidth]{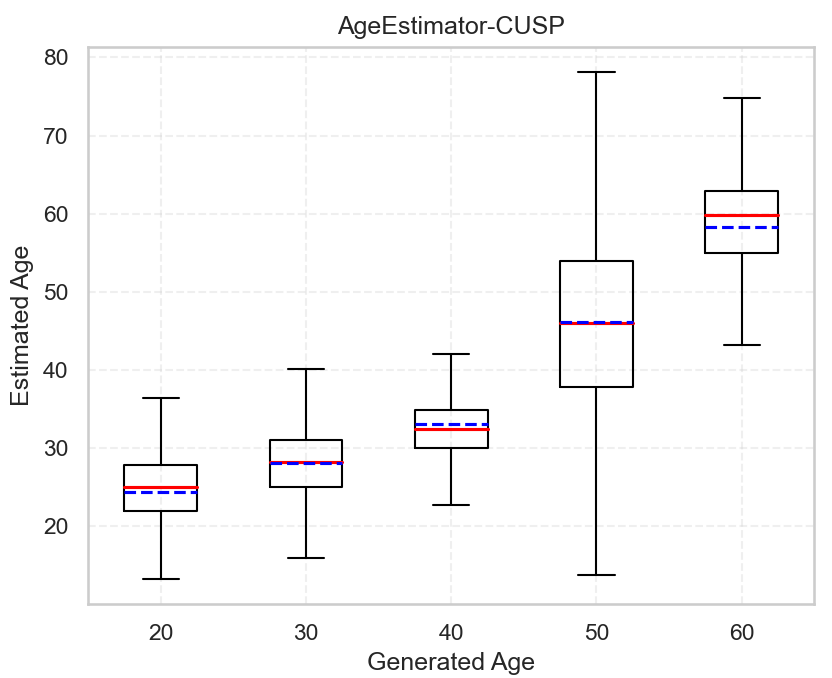}
    \includegraphics[width=0.33\linewidth]{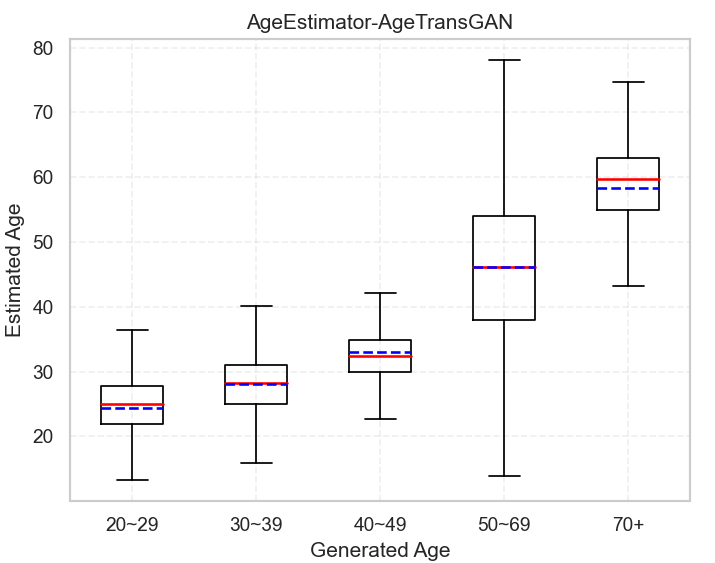}\\
    (a)\qquad \qquad \qquad \qquad \qquad \qquad \qquad \qquad (b) \qquad \qquad \qquad \qquad \qquad \qquad \qquad \qquad  (c) \\
    \includegraphics[width=0.32\linewidth]{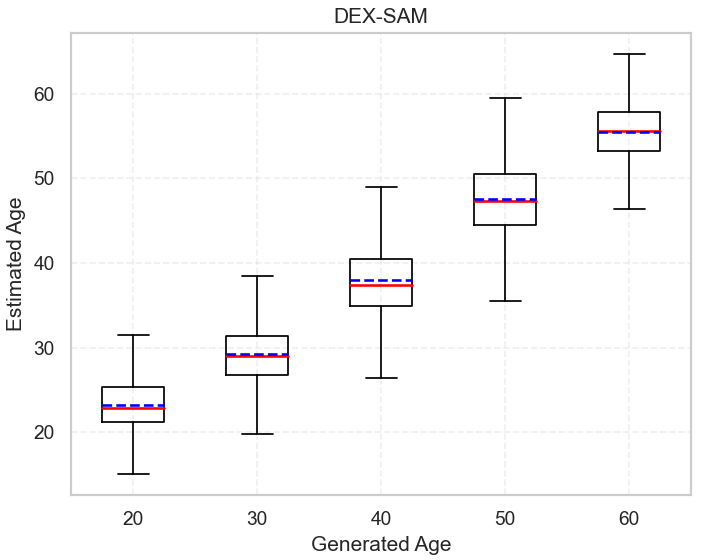}    
    \includegraphics[width=0.32\linewidth]{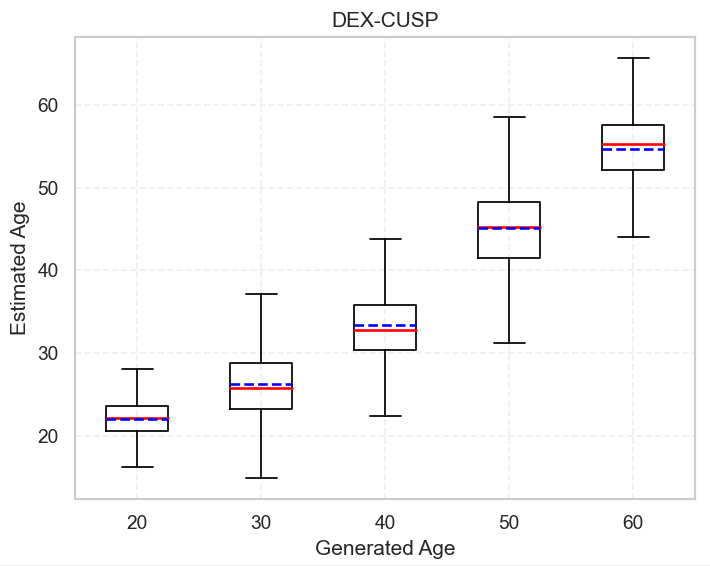}
    \includegraphics[width=0.32\linewidth]{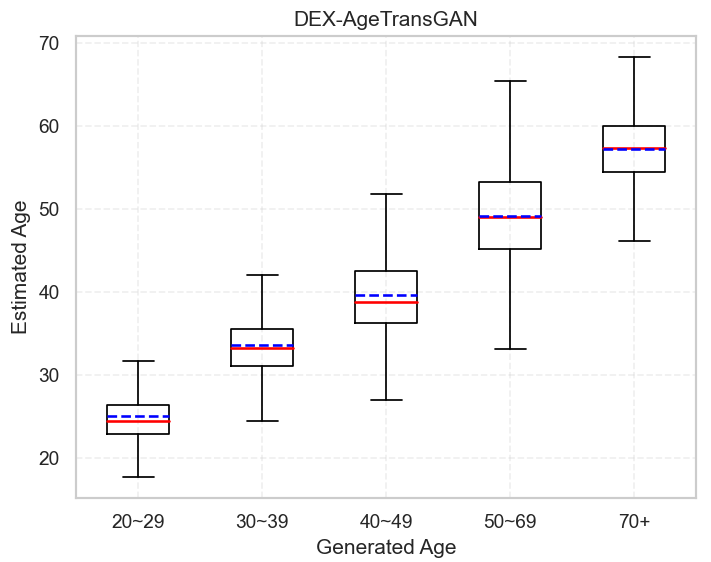}\\
    (d)\qquad \qquad \qquad \qquad \qquad \qquad \qquad \qquad (e) \qquad \qquad \qquad \qquad \qquad \qquad \qquad \qquad  (f) \\
    \caption{Ageing accuracy analysis by using two age estimators. The first row uses the estimator proposed by Hsu et al.~\cite{10.1007/978-3-031-19775-8_34}. The second row adopts DEX~\cite{Rothe_2015_ICCV_Workshops} estimator. (a) and (d) show the estimated results of samples generated by the SAM method. (b) and (e) show the estimated results of samples generated by the CUSP method. (c) and (f) show the estimated results of samples generated by the AgeTransGAN method.}
    \label{estimator}
\end{figure*}

In order to evaluate the identity preservation of the synthetic faces, we have designed two analysis identity similarity experiments as demonstrated in Figure~\ref{process_id}. Motivated by the concept of consistency loss in CycleGAN~\cite{Zhu_2017_ICCV}, we introduce a novel approach to assess the identity preservation of synthetic age images through self-consistency, which is the first application of such an assessment method. The first similarity shows the similarity between the synthesized images of different ages and the original images. The second similarity shows the reverse similarity between the synthesized reverse images and the original images. In this study, we assess the ability of synthetic ageing methods to preserve identity by comparing the match scores derived from two distinct analyses. This comparative evaluation allows us to gauge the overall efficacy of these methods in maintaining identity across age-transformed facial images. 

Figure~\ref{id_ana} exhibits the experimental results of identity preservation. Figure~\ref{id_ana} (a)(b)(c) shows the match score distribution of synthetic data with various age ranges. As the age of the synthesized image increases, the score of the genuine distribution becomes progressively smaller. The impostor distribution remains almost unchanged. From Figure~\ref{id_ana} (a)(b)(c), we find that the identity of AgeTransGAN is maintained better for generating images of different age ranges. Figure~\ref{id_ana} (d)(e)(f) shows the match score distribution of synthetic data that reverse back from various age ranges. For the different synthetic ageing algorithms, the genuine distribution scores decreased to varying degrees during the process of reverse back to the 20s. The distribution of impostors exhibits minimal alteration during this process. As it can be observed from Figure~\ref{id_ana} (d)(e)(f), the identity similarity of the reverse back ability of SAM with AgeTransGAN is more stable than CUSP. From this experiment, we found that the AgeTransGAN has better identity preservation capability than the other two algorithms.

\subsection{Age Accuracy}
In this experiment, the CelebA-HQ dataset is used to generate exact synthetic ageing data including $20-, 30-, 40-, 50-, 60-$year-old faces by using CUSP and SAM methods. We have also generated specific age brackets such that 20-29, 30-39, 40-49, 50-69, and 70+ years old faces by using the AgeTransGAN method, as AgeTransGAN can only generate images for different age group ranges. In the next stage two pretrained age estimators are used to estimate the synthetic faces, which shows the accuracy of the synthetic facial aging samples. The results are shown in Figure~\ref{estimator}. As depicted in Figure~\ref{estimator} (a)(d), the outcomes of age estimators for the 20-year-old image synthesized by the SAM algorithm indicated a slight overestimation of age, appearing slightly older than 20 years. Conversely, the age detection results for the 60-year-old image synthesized by the SAM algorithm suggested a slight underestimation of age, indicating an age slightly less than 60 years. However, the remaining assessments of synthesized image ages closely approximated the actual ages. As illustrated in Figure~\ref{estimator} (b)(e), the age detection outcomes for the 20-year-old image synthesized by the CUSP algorithm exhibited slightly older than 20 years old. The age detection results for the 40-year-old image synthesized by the CUSP algorithm showed less than 40 years old. The rest assessments of synthesized image ages are relatively close to the real situation. Figure~\ref{estimator} (c)(f) shows the age estimators’ results of the images synthesized by the AgeTransGAN algorithm for the age group of 40 years and above less than the target age range. Figure~\ref{estimator} shows that the SAM algorithm has performed slightly better than the remaining two algorithms in terms of age accuracy.

\begin{figure}[thpb]
    \centering
    \includegraphics[width=\linewidth]{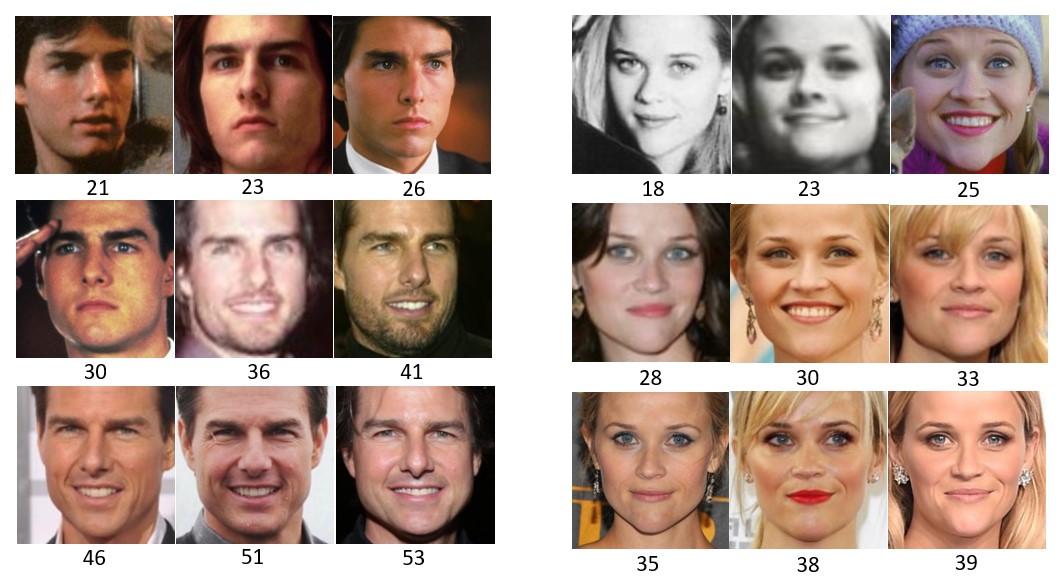}
    \caption{An example of subjects at different ages.}
    \label{sample}
\end{figure}
\section{The robustness of FR on real-world vs. synthetic ageing data}\label{robustness}
In this section, we discuss the effect of synthetic and real face ageing on the face recognition algorithm, which helps us to understand the effect of the age gap for real-world data and synthetic ageing data on the state-of-the-art deep learning-based face recognition model. In this work, we have used the ArcFace model, which is counted as one of the state-of-the-art face recognition models. This is because our previous research work~\cite{10242105} on the effect of age group in face authentication showed that using different deep learning-based face recognition models only differed in accuracy and the trend of the ROC curves was similar. 

\subsection{Real-World Ageing Effect}

This experiment was conducted to quantify the effect of real ageing intervals on the face recognition model. We have used the B3FD dataset and generated different age pairs by different age intervals. Then we tested these age pairs from the real-world dataset on the ArcFace model. Figure~\ref{sample} shows an example of subjects that changed after various years from the B3FD dataset. 

\begin{figure}[!thpb]
    \centering
    \includegraphics[width=0.98\linewidth]{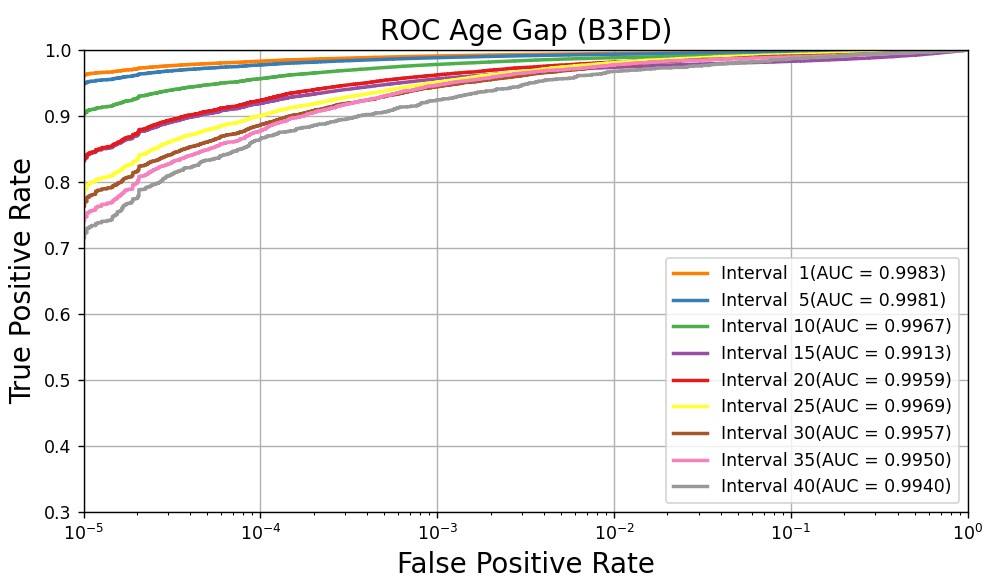}\\(a)\\
    \includegraphics[width=0.98\linewidth]{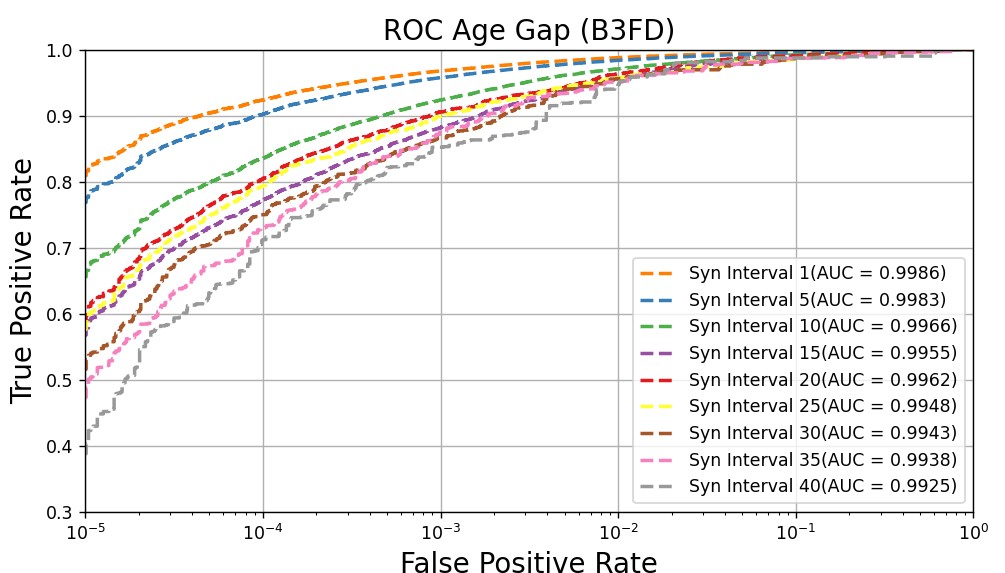}\\(b)\\
    \caption{Ageing Effect. (a) real-world. (b) synthetic agine effect.}
    \label{effects}
\end{figure}

In this work, we have used the Receiver Operating Characteristic (ROC) curves to quantify the ageing effect on the performance of the FR model. The ROC curve can visually demonstrate the uniqueness of the identity and is widely used to evaluate the performance of face recognition models~\cite{9165737}. Figure~\ref{effects} (a) shows the ROC curves for various age intervals from the B3FD dataset tested on the ArcFace model. As it can be seen in Figure~\ref{effects} (a), as the age interval widens, the effectiveness of the face recognition system decreases. Age intervals spanning up to 5 years exhibit a lesser impact on the performance of the face recognition system.

\subsection{Synthetic Ageing Effect}

In this experiment, we have used the synthetic ageing method, SAM, to generate synthetic ageing samples. Because SAM can generate images of the target age, which makes our experimental setup more flexible. Then we quantify the effect of ageing on the FR model by using these synthetic aging samples. The same image pairs have been adopted in this process. The only difference is that we have used the image, which is a small age in each image pair, and using this image to generate the same age of the other image. In this way, the age interval between the real-world image and generated image will keep unchanged. After that, we quantify these image pairs by using the ArcFace model and calculate the ROC curves as shown in Figure~\ref{effects} (b). In Figure~\ref{effects} (b), we observed a decrease in the effectiveness of the face recognition system as the age interval widens. The phenomenon is consistent with the ROC curves of the real-world age intervals in Figure~\ref{effects} (a). The difference is that the impact of age intervals derived from synthesized images on the face recognizer is marginally less pronounced compared to the effect exerted by age intervals from real images on the face recognizer. Our findings suggest that synthetic age images exhibit certain deficiencies when compared to real age images, which is a reasonable phenomenon given the complexities involved in synthesizing age transformations accurately.

\section{Facilitating age-invariant FR with synthetic ageing data}\label{finetune}

In this section, we aim to investigate the feasibility of leveraging synthetic ageing data to enhance the robustness of the face recognition algorithm. Keeping this in view, various networks were trained by adopting different training datasets to illustrate how synthetic ageing data facilitates the age-invariant face recognition model and if there a specific synthetic age samples that get a better result on the improvement of the face recognition model.  

\begin{table*}[!htpb]
\centering
\caption{Experimental setting for tuning models. }
\label{tab:setting}
\resizebox{\textwidth}{!}{%
\begin{tabular}{|c|l|c|}
\hline
Model name & Training dataset used in finetuning the face recognition model & Number of training data \\ \hline
Baseline      & CelebA-HQ dataset                                             & 28690 \\ \hline
CelebA-Syn-25 & Synthetic samples that are 25 years old and CelebA-HQ dataset & 57588 \\ \hline
CelebA-Syn-35 & Synthetic samples that are 35 years old and CelebA-HQ dataset & 57588 \\ \hline
CelebA-Syn-45 & Synthetic samples that are 45 years old and CelebA-HQ dataset & 57581 \\ \hline
CelebA-Syn-55 & Synthetic samples that are 55 years old and CelebA-HQ dataset & 57578 \\ \hline
CelebA-Syn-65 & Synthetic samples that are 65 years old and CelebA-HQ dataset & 57567 \\ \hline
\end{tabular}%
}
\end{table*}

\begin{table*}[!thpb]
\centering
\caption{True positive rates (\%) with a false match rate of $10^{-3}$ on the original B3FD dataset.}
\label{tab:result}
\resizebox{\textwidth}{!}{%
\begin{tabular}{|c|c|c|c|c|c|c|}
\hline
Model         & Age Gap 1 & Age Gap 5 & Age Gap 10 & Age Gap 20 & Age Gap 30 & Age Gap 40 \\ \hline
Baseline      & 98.34     & 97.21     & 94.86      & 94.15      & 90.54      & 88.88      \\ \hline
CelebA-Syn-25 & 98.76     & 97.86     & 96.46      & 94.68      & 91.98      & 90.82      \\ \hline
CelebA-Syn-35 & 98.87     & 98.20     & 96.77      & 95.47      & 93.02      & 91.21      \\ \hline
CelebA-Syn-45 & 98.87     & 98.12     & 96.58      & 95.22      & 92.77      & 90.49      \\ \hline
CelebA-Syn-55 & 99.06     & 98.42     & 97.27      & 95.90      & 93.29      & 92.21      \\ \hline
CelebA-Syn-65 & 98.95     & 98.24     & 97.02      & 95.55      & 92.48      & 90.99      \\ \hline
\end{tabular}%
}
\end{table*}

In this experiment, CelebA-HQ and synthetic ageing samples generated from the CelebA-HQ dataset by the SAM method are chosen as the training datasets. The validation datasets are using lfw, agedb\_30 and cfp\_fp. After finetuning, we tested the model on a real-world B3FD dataset, which is one of the largest ageing datasets. The different experimental settings for training the models are as follows in Table~\ref{tab:setting}. The pre-trained model~\footnote{Pretrained model:} has been used as the original model to save training time.
Different training datasets are used in tuning the face recognition model, then the tuned model is adopted to test the effect of the various age gaps, which are 1 year, 5 years, 10 years, 20 years, 30 years and 40 years. 

The experimental results are shown in Table~\ref{tab:result}. From Table~\ref{tab:result}, we can conclude that the baseline model has the lowest accuracy, while the model finetuned with synthetic ageing data has shown better results for the age gap. This means that synthetic ageing data could be utilized as a beneficial resource to improve the effectiveness of face recognition. Moreover, we can find \textit{CelebA-Syn-55} model has the best results on a large age gap, which shows face recognition models fine-tuned using synthetic data of different ages have different robustness to age intervals.
\section{Conclusion}\label{conclusion}
In this study, we have devised an experimental framework aimed at assessing the efficacy of synthetic ageing samples and investigating their impact on a large-scale real-world face ageing dataset. Moreover, we have quantified the potential usage of synthetic face ageing data as an argument method to improve the robustness of face recognition on large age gaps. The overall experimental analysis yields several noteworthy findings. Notably, despite the advancements in synthetic age algorithms such as SAM, CUSP, and AgeTransGAN, these methods still exhibit limitations in preserving facial identity when compared to real world images. Moreover, irrespective of whether synthetic or real age images are employed, the performance of face recognition models declines with increasing age gaps. However, as the final research outcome we concluded that, leveraging synthetic age images for training purposes demonstrates robust results for face recognizer algorithms against the age gap promise.  

As a prospect for future research, we intend to conduct a comprehensive comparative analysis  of different synthetic ageing samples by using subjective evaluation methodologies. Furthermore, we aim to expand our assessment of synthetic aging data and devise metrics to quantify the efficacy of various synthetic aging techniques.  
\bibliographystyle{IEEEtran}
\bibliography{ref.bib}

\begin{thebibliography}{10}
\providecommand{\url}[1]{#1}
\csname url@samestyle\endcsname
\providecommand{\newblock}{\relax}
\providecommand{\bibinfo}[2]{#2}
\providecommand{\BIBentrySTDinterwordspacing}{\spaceskip=0pt\relax}
\providecommand{\BIBentryALTinterwordstretchfactor}{4}
\providecommand{\BIBentryALTinterwordspacing}{\spaceskip=\fontdimen2\font plus
\BIBentryALTinterwordstretchfactor\fontdimen3\font minus \fontdimen4\font\relax}
\providecommand{\BIBforeignlanguage}[2]{{%
\expandafter\ifx\csname l@#1\endcsname\relax
\typeout{** WARNING: IEEEtran.bst: No hyphenation pattern has been}%
\typeout{** loaded for the language `#1'. Using the pattern for}%
\typeout{** the default language instead.}%
\else
\language=\csname l@#1\endcsname
\fi
#2}}
\providecommand{\BIBdecl}{\relax}
\BIBdecl

\bibitem{9025674}
N.~Srinivas, K.~Ricanek, D.~Michalski, D.~S. Bolme, and M.~King, ``Face recognition algorithm bias: Performance differences on images of children and adults,'' in \emph{2019 IEEE/CVF Conference on Computer Vision and Pattern Recognition Workshops (CVPRW)}, 2019, pp. 2269--2277.

\bibitem{7815403}
L.~Best-Rowden and A.~K. Jain, ``Longitudinal study of automatic face recognition,'' \emph{IEEE Transactions on Pattern Analysis and Machine Intelligence}, vol.~40, no.~1, pp. 148--162, 2018.

\bibitem{10121472}
Y.~Zhang, T.~Wang, R.~Zhao, W.~Wen, and Y.~Zhu, ``Rapp: Reversible privacy preservation for various face attributes,'' \emph{IEEE Transactions on Information Forensics and Security}, vol.~18, pp. 3074--3087, 2023.

\bibitem{993553}
A.~Lanitis, C.~Taylor, and T.~Cootes, ``Toward automatic simulation of aging effects on face images,'' \emph{IEEE Transactions on Pattern Analysis and Machine Intelligence}, vol.~24, no.~4, pp. 442--455, 2002.

\bibitem{8014984}
S.~Moschoglou, A.~Papaioannou, C.~Sagonas, J.~Deng, I.~Kotsia, and S.~Zafeiriou, ``Agedb: The first manually collected, in-the-wild age database,'' in \emph{2017 IEEE Conference on Computer Vision and Pattern Recognition Workshops (CVPRW)}, 2017, pp. 1997--2005.

\bibitem{1613043}
K.~Ricanek and T.~Tesafaye, ``Morph: a longitudinal image database of normal adult age-progression,'' in \emph{7th International Conference on Automatic Face and Gesture Recognition (FGR06)}, 2006, pp. 341--345.

\bibitem{10.1007/978-3-030-58539-6_44}
R.~Or-El, S.~Sengupta, O.~Fried, E.~Shechtman, and I.~Kemelmacher-Shlizerman, ``Lifespan age transformation synthesis,'' in \emph{Computer Vision -- ECCV 2020}, A.~Vedaldi, H.~Bischof, T.~Brox, and J.-M. Frahm, Eds.\hskip 1em plus 0.5em minus 0.4em\relax Cham: Springer International Publishing, 2020, pp. 739--755.

\bibitem{10.1007/978-3-031-19775-8_34}
G.-S. Hsu, R.-C. Xie, Z.-T. Chen, and Y.-H. Lin, ``Agetransgan for facial age transformation with rectified performance metrics,'' in \emph{Computer Vision -- ECCV 2022}, S.~Avidan, G.~Brostow, M.~Ciss{\'e}, G.~M. Farinella, and T.~Hassner, Eds.\hskip 1em plus 0.5em minus 0.4em\relax Cham: Springer Nature Switzerland, 2022, pp. 580--595.

\bibitem{10.1145/3450626.3459805}
\BIBentryALTinterwordspacing
Y.~Alaluf, O.~Patashnik, and D.~Cohen-Or, ``Only a matter of style: age transformation using a style-based regression model,'' \emph{ACM Trans. Graph.}, vol.~40, no.~4, jul 2021. [Online]. Available: \url{https://doi.org/10.1145/3450626.3459805}
\BIBentrySTDinterwordspacing

\bibitem{10.1007/978-3-031-19787-1_32}
G.~Gomez-Trenado, S.~Lathuili{\`e}re, P.~Mesejo, and {\'O}.~Cord{\'o}n, ``Custom structure preservation in face aging,'' in \emph{Computer Vision -- ECCV 2022}, S.~Avidan, G.~Brostow, M.~Ciss{\'e}, G.~M. Farinella, and T.~Hassner, Eds.\hskip 1em plus 0.5em minus 0.4em\relax Cham: Springer Nature Switzerland, 2022, pp. 565--580.

\bibitem{6327355}
B.~F. Klare, M.~J. Burge, J.~C. Klontz, R.~W. Vorder~Bruegge, and A.~K. Jain, ``Face recognition performance: Role of demographic information,'' \emph{IEEE Transactions on Information Forensics and Security}, vol.~7, no.~6, pp. 1789--1801, 2012.

\bibitem{9093357}
V.~Albiero, K.~W. Bowyer, K.~Vangara, and M.~C. King, ``Does face recognition accuracy get better with age? deep face matchers say no,'' in \emph{2020 IEEE Winter Conference on Applications of Computer Vision (WACV)}, 2020, pp. 250--258.

\bibitem{WU2019116}
\BIBentryALTinterwordspacing
S.~wu and D.~Wang, ``Effect of subject's age and gender on face recognition results,'' \emph{Journal of Visual Communication and Image Representation}, vol.~60, pp. 116--122, 2019. [Online]. Available: \url{https://www.sciencedirect.com/science/article/pii/S1047320319300197}
\BIBentrySTDinterwordspacing

\bibitem{4409069}
H.~Ling, S.~Soatto, N.~Ramanathan, and D.~W. Jacobs, ``A study of face recognition as people age,'' in \emph{2007 IEEE 11th International Conference on Computer Vision}, 2007, pp. 1--8.

\bibitem{8014816}
\BIBentryALTinterwordspacing
D.~Deb, L.~Best-Rowden, and A.~K. Jain, ``Face recognition performance under aging,'' in \emph{2017 IEEE Conference on Computer Vision and Pattern Recognition Workshops (CVPRW)}.\hskip 1em plus 0.5em minus 0.4em\relax Los Alamitos, CA, USA: IEEE Computer Society, jul 2017, pp. 548--556. [Online]. Available: \url{https://doi.ieeecomputersociety.org/10.1109/CVPRW.2017.82}
\BIBentrySTDinterwordspacing

\bibitem{sawant2019age}
M.~M. Sawant and K.~M. Bhurchandi, ``Age invariant face recognition: a survey on facial aging databases, techniques and effect of aging,'' \emph{Artificial Intelligence Review}, vol.~52, pp. 981--1008, 2019.

\bibitem{9897043}
M.~Grimmer, H.~Zhang, R.~Ramachandra, K.~Raja, and C.~Busch, ``Time flies by: Analyzing the impact of face ageing on the recognition performance with synthetic data,'' in \emph{2022 International Conference of the Biometrics Special Interest Group (BIOSIG)}, 2022, pp. 1--6.

\bibitem{9157070}
Y.~Shen, J.~Gu, X.~Tang, and B.~Zhou, ``Interpreting the latent space of gans for semantic face editing,'' in \emph{2020 IEEE/CVF Conference on Computer Vision and Pattern Recognition (CVPR)}, 2020, pp. 9240--9249.

\bibitem{6751468}
D.~Gong, Z.~Li, D.~Lin, J.~Liu, and X.~Tang, ``Hidden factor analysis for age invariant face recognition,'' in \emph{2013 IEEE International Conference on Computer Vision}, 2013, pp. 2872--2879.

\bibitem{8323422}
H.~Li, H.~Hu, and C.~Yip, ``Age-related factor guided joint task modeling convolutional neural network for cross-age face recognition,'' \emph{IEEE Transactions on Information Forensics and Security}, vol.~13, no.~9, pp. 2383--2392, 2018.

\bibitem{9711408}
X.~Hou, Y.~Li, and S.~Wang, ``Disentangled representation for age-invariant face recognition: A mutual information minimization perspective,'' in \emph{2021 IEEE/CVF International Conference on Computer Vision (ICCV)}, 2021, pp. 3672--3681.

\bibitem{10.1145/3472810}
\BIBentryALTinterwordspacing
C.~Yan, L.~Meng, L.~Li, J.~Zhang, Z.~Wang, J.~Yin, J.~Zhang, Y.~Sun, and B.~Zheng, ``Age-invariant face recognition by multi-feature fusionand decomposition with self-attention,'' \emph{ACM Trans. Multimedia Comput. Commun. Appl.}, vol.~18, no.~1s, jan 2022. [Online]. Available: \url{https://doi.org/10.1145/3472810}
\BIBentrySTDinterwordspacing

\bibitem{9146699}
J.~Zhao, S.~Yan, and J.~Feng, ``Towards age-invariant face recognition,'' \emph{IEEE Transactions on Pattern Analysis and Machine Intelligence}, vol.~44, no.~1, pp. 474--487, 2022.

\bibitem{8936470}
L.~Du, H.~Hu, and Y.~Wu, ``Cycle age-adversarial model based on identity preserving network and transfer learning for cross-age face recognition,'' \emph{IEEE Transactions on Information Forensics and Security}, vol.~15, pp. 2241--2252, 2020.

\bibitem{9931965}
Z.~Huang, J.~Zhang, and H.~Shan, ``When age-invariant face recognition meets face age synthesis: A multi-task learning framework and a new benchmark,'' \emph{IEEE Transactions on Pattern Analysis and Machine Intelligence}, vol.~45, no.~6, pp. 7917--7932, 2023.

\bibitem{electronics12112369}
\BIBentryALTinterwordspacing
S.~Li and H.~J. Lee, ``Gfam: A gender-preserving face aging model for age imbalance data,'' \emph{Electronics}, vol.~12, no.~11, 2023. [Online]. Available: \url{https://www.mdpi.com/2079-9292/12/11/2369}
\BIBentrySTDinterwordspacing

\bibitem{9412383}
X.~Yao, G.~Puy, A.~Newson, Y.~Gousseau, and P.~Hellier, ``High resolution face age editing,'' in \emph{2020 25th International Conference on Pattern Recognition (ICPR)}, 2021, pp. 8624--8631.

\bibitem{chen2023face}
X.~Chen and S.~Lathuili{\`e}re, ``Face aging via diffusion-based editing,'' \emph{arXiv preprint arXiv:2309.11321}, 2023.

\bibitem{liu2015faceattributes}
Z.~Liu, P.~Luo, X.~Wang, and X.~Tang, ``Deep learning face attributes in the wild,'' in \emph{Proceedings of International Conference on Computer Vision (ICCV)}, December 2015.

\bibitem{bevsenic2022picking}
K.~Be{\v{s}}eni{\'c}, J.~Ahlberg, and I.~S. Pand{\v{z}}i{\'c}, ``Picking out the bad apples: unsupervised biometric data filtering for refined age estimation,'' \emph{The Visual Computer}, pp. 1--19, 2022.

\bibitem{10.1007/978-3-319-10599-4_49}
B.-C. Chen, C.-S. Chen, and W.~H. Hsu, ``Cross-age reference coding for age-invariant face recognition and retrieval,'' in \emph{Computer Vision -- ECCV 2014}, D.~Fleet, T.~Pajdla, B.~Schiele, and T.~Tuytelaars, Eds.\hskip 1em plus 0.5em minus 0.4em\relax Cham: Springer International Publishing, 2014, pp. 768--783.

\bibitem{Rothe2016DeepEO}
\BIBentryALTinterwordspacing
R.~Rothe, R.~Timofte, and L.~V. Gool, ``Deep expectation of real and apparent age from a single image without facial landmarks,'' \emph{International Journal of Computer Vision}, vol. 126, pp. 144 -- 157, 2016. [Online]. Available: \url{https://api.semanticscholar.org/CorpusID:207252421}
\BIBentrySTDinterwordspacing

\bibitem{Liu_2019_CVPR}
Y.~Liu, Q.~Li, and Z.~Sun, ``Attribute-aware face aging with wavelet-based generative adversarial networks,'' in \emph{Proceedings of the IEEE/CVF Conference on Computer Vision and Pattern Recognition (CVPR)}, June 2019.

\bibitem{Rothe_2015_ICCV_Workshops}
R.~Rothe, R.~Timofte, and L.~Van~Gool, ``Dex: Deep expectation of apparent age from a single image,'' in \emph{Proceedings of the IEEE International Conference on Computer Vision (ICCV) Workshops}, December 2015.

\bibitem{Zhu_2017_ICCV}
J.-Y. Zhu, T.~Park, P.~Isola, and A.~A. Efros, ``Unpaired image-to-image translation using cycle-consistent adversarial networks,'' in \emph{Proceedings of the IEEE International Conference on Computer Vision (ICCV)}, Oct 2017.

\bibitem{chen2023topiq}
C.~Chen, J.~Mo, J.~Hou, H.~Wu, L.~Liao, W.~Sun, Q.~Yan, and W.~Lin, ``Topiq: A top-down approach from semantics to distortions for image quality assessment,'' \emph{arXiv preprint arXiv:2308.03060}, 2023.

\bibitem{mittal2012no}
A.~Mittal, A.~K. Moorthy, and A.~C. Bovik, ``No-reference image quality assessment in the spatial domain,'' \emph{IEEE Transactions on image processing}, vol.~21, no.~12, pp. 4695--4708, 2012.

\bibitem{10242105}
W.~Yao, M.~A. Farooq, J.~Lemley, and P.~Corcoran, ``A study on the effect of ageing in facial authentication and the utility of data augmentation to reduce performance bias across age groups,'' \emph{IEEE Access}, vol.~11, pp. 97\,118--97\,134, 2023.

\bibitem{9165737}
V.~Varkarakis, S.~Bazrafkan, G.~Costache, and P.~Corcoran, ``Validating seed data samples for synthetic identities – methodology and uniqueness metrics,'' \emph{IEEE Access}, vol.~8, pp. 152\,532--152\,550, 2020.

\bibitem{7553523}
K.~Zhang, Z.~Zhang, Z.~Li, and Y.~Qiao, ``Joint face detection and alignment using multitask cascaded convolutional networks,'' \emph{IEEE Signal Processing Letters}, vol.~23, no.~10, pp. 1499--1503, 2016.

\bibitem{9201174}
L.~Du and H.~Hu, ``Cross-age identity difference analysis model based on image pairs for age invariant face verification,'' \emph{IEEE Transactions on Circuits and Systems for Video Technology}, vol.~31, no.~7, pp. 2675--2685, 2021.

\bibitem{Wang_2019_CVPR}
H.~Wang, D.~Gong, Z.~Li, and W.~Liu, ``Decorrelated adversarial learning for age-invariant face recognition,'' in \emph{Proceedings of the IEEE/CVF Conference on Computer Vision and Pattern Recognition (CVPR)}, June 2019.

\bibitem{Huang_2021_CVPR}
Z.~Huang, J.~Zhang, and H.~Shan, ``When age-invariant face recognition meets face age synthesis: A multi-task learning framework,'' in \emph{Proceedings of the IEEE/CVF Conference on Computer Vision and Pattern Recognition (CVPR)}, June 2021, pp. 7282--7291.

\bibitem{zhang2023age}
Z.~Zhang, S.~Yin, and L.~Cao, ``Age-invariant face recognition based on identity-age shared features,'' \emph{The Visual Computer}, pp. 1--10, 2023.

\bibitem{DBLP:journals/corr/abs-1708-08197}
\BIBentryALTinterwordspacing
T.~Zheng, W.~Deng, and J.~Hu, ``Cross-age {LFW:} {A} database for studying cross-age face recognition in unconstrained environments,'' \emph{CoRR}, vol. abs/1708.08197, 2017. [Online]. Available: \url{http://arxiv.org/abs/1708.08197}
\BIBentrySTDinterwordspacing

\bibitem{LFWTech}
G.~B. Huang, M.~Ramesh, T.~Berg, and E.~Learned-Miller, ``Labeled faces in the wild: A database for studying face recognition in unconstrained environments,'' University of Massachusetts, Amherst, Tech. Rep. 07-49, October 2007.

\end{thebibliography}











\newpage

\vfill

\end{document}